\documentclass[final]{cvpr}
\usepackage{comment}
\usepackage{times}
\usepackage{graphicx}
\usepackage{amsmath}
\usepackage{amssymb}
\usepackage{makecell}
\usepackage{verbatim}
\usepackage{caption}
\usepackage{booktabs}
\usepackage{url}
\usepackage{ragged2e}
\usepackage{amsmath,amssymb}
\usepackage{utfsym}
\usepackage[ruled,linesnumbered, vlined]{algorithm2e}
\usepackage[pagebackref,breaklinks,colorlinks]{hyperref}

\title{DiffMAC: Diffusion \underline{M}anifold H\underline{a}llucination \underline{C}orrection for High Generalization Blind Face Restoration}

\author{
    Nan Gao$^{1}$, Jia Li$^{2}$, Huaibo Huang$^{1}$, Zhi Zeng$^{3}$, Ke Shang$^{2}$, Shuwu Zhang$^{3}$, Ran He$^{1}$\\
    $^1$ Institute of Automation Chinese Academy of Sciences, Beijing, China\\
    $^2$ PCIE, Lenovo Research, Beijing, China\\
    $^3$ Beijing University of Posts and Telecommunications\\
}

\begin{document}


\maketitle
\begin{abstract}
Blind face restoration (BFR) is a highly challenging problem due to the uncertainty of degradation patterns. Current methods have low generalization across photorealistic and heterogeneous domains. In this paper, we propose a Diffusion-Information-Diffusion (DID) framework to tackle diffusion manifold hallucination correction (DiffMAC), which achieves high-generalization face restoration in diverse degraded scenes and heterogeneous domains. Specifically, the first diffusion stage aligns the restored face with spatial feature embedding of the low-quality face based on AdaIN, which synthesizes degradation-removal results but with uncontrollable artifacts for some hard cases. Based on Stage I, Stage II considers information compression using manifold information bottleneck (MIB) and finetunes the first diffusion model to improve facial fidelity. DiffMAC effectively fights against blind degradation patterns and synthesizes high-quality faces with attribute and identity consistencies. Experimental results demonstrate the superiority of DiffMAC over state-of-the-art methods, with a high degree of generalization in real-world and heterogeneous settings. The source code and models will be public.
\end{abstract}

\section{Introduction}

\begin{figure}[h]
  \centering
  
  \includegraphics[width=0.73\linewidth]{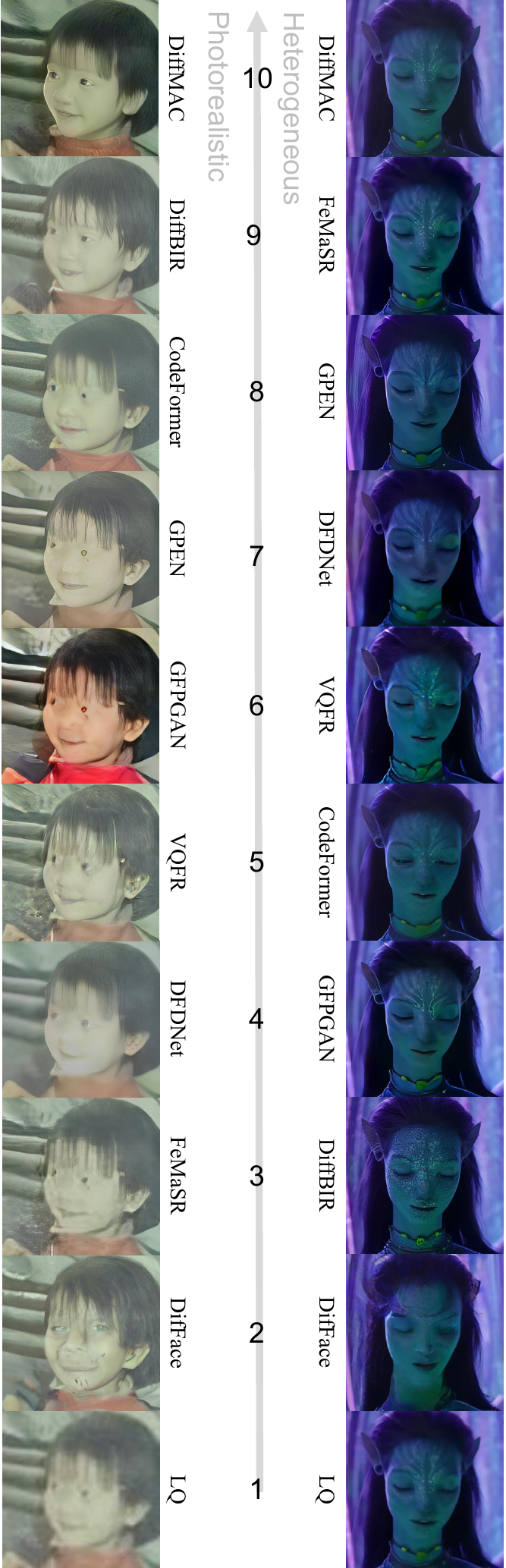}
   \caption{Our proposed DiffMAC approach achieves more promising BFR results with higher fidelity both in photorealistic and heterogeneous scenarios. More comparisons are illustrated in Fig \ref{fig:stage2}, \ref{fig:photo}, \ref{fig:heter} and \ref{fig:heterm}.}
   \label{fig:m2}
\end{figure}

High Generalization blind face restoration (HG-BFR) is a challenging task, particularly when dealing with severe degradation factors and out-of-domain heterogeneous scenes (Fig \ref{fig:m2}). HG-BFR not only means facilitating restoration quality to a high photorealistic level, but also considering scene attributes (e.g., heterogeneous style, original cloth textures, facial appearance, or identity). Appropriate reinforcements based on original content are expected, while a large visual gap caused by attribute and identity inconsistencies will result in a negative subjective perception. 
\begin{figure}[tbp]
  \centering
  
  \includegraphics[width=0.66\linewidth]{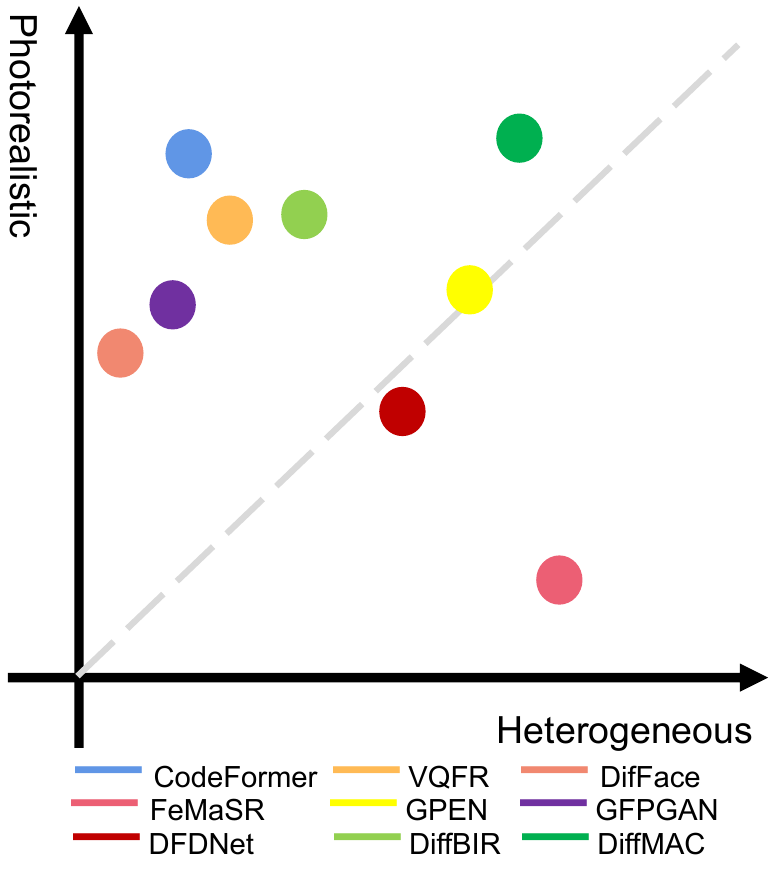}
   \caption{Our proposed DiffMAC approach achieves less face hallucination brought from deep learning model and possesses better visual performance, compared with other state-of-the-art (CodeFormer \cite{codeformer}, VQFR \cite{vqfr}, DifFace \cite{difface}, FeMaSR \cite{femasr}, GPEN \cite{GPEN}, GFPGAN \cite{gfp2021}, DFDNet \cite{dfdnet2020}, DiffBIR \cite{diffbir}) in photorealistic and heterogeneous domains. The subjective study refers to Table \ref{tab:p2h}.}
   \label{fig:motivation}
\end{figure}

Recently, GAN-based methods \cite{vqfr, GPEN, gfp2021, femasr, dfdnet2020} and Transformer-based methods \cite{codeformer, difface, diffbir} have achieved acknowledged BFR performance, which leverage high feature representation ability of pre-trained models, such as VQ-GAN \cite{vqgan}, StyleGAN \cite{style2019}, stable diffusion \cite{latentdiff}. As shown in Fig \ref{fig:motivation}, most of these methods deal with BFR in photorealistic domains well, while not capable of high fidelity BFR in heterogeneous domains. One essential reason is the training dataset FFHQ \cite{style2019} mainly contains faces in the real world, which is easy to lead to a prior bias away from the heterogeneous face domain. Specifically, as shown in \ref{fig:m2}, there are color distortion \cite{dfdnet2020}, texture distortion \cite{vqfr, GPEN, gfp2021}, expression hallucination \cite{codeformer}, identity distortion \cite{difface}, mesh artifacts \cite{diffbir}.

Furthermore, for some faces with severe degradation, the restored results have heavy hallucination with facial distortion in real-world domain, e.g., heavy color distortion \cite{gfp2021}, texture distortion \cite{vqfr, femasr, difface}, mode collapse of eyes\cite{codeformer, GPEN}, identity distortion \cite{difface}, unsharp edges\cite{dfdnet2020,diffbir}. Our approach DiffMAC involves information compression learning \cite{iba} for the manifold of diffusion model \cite{latentdiff} to implement photographic BFR in diverse degradation levels (Fig \ref{fig:stage2}, Fig \ref{fig:photo}), which is also promising for heterogeneous BFR (Fig \ref{fig:heter}, Fig \ref{fig:heterm}).

\begin{figure}[htbp]
  \centering
  
  \includegraphics[width=1\linewidth]{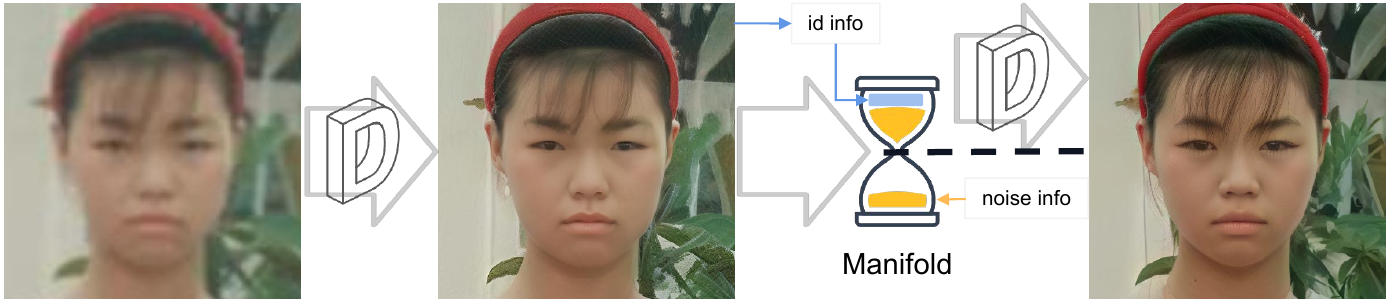}
  \includegraphics[width=1\linewidth]{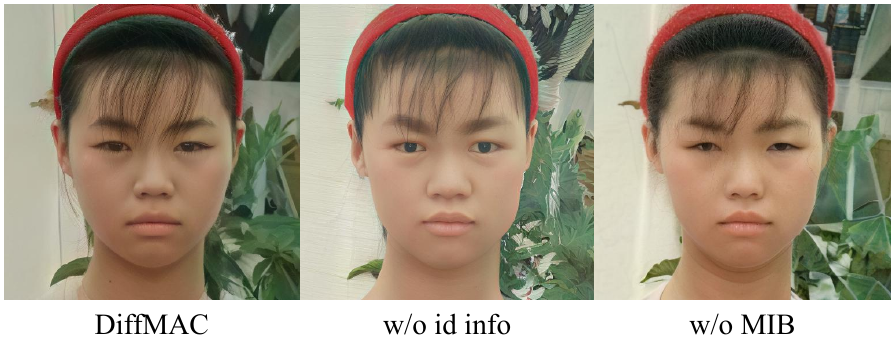}
   \caption{DID (diffusion-information-diffusion) pipeline of our proposed DiffMAC where the first diffusion model transfers random degraded patterns to a relatively stable restoration mode that is friendly to conduct information bottleneck. Another controllable diffusion model based on the compressed manifold is then leveraged to obtain better BFR. More results are shown in Fig \ref{fig:stage2}.}
   \label{fig:m3}
\end{figure}

Diffusion model has been potentially used on high-quality image syntheses \cite{latentdiff, controlnet, diffbir}. However, BIR in the severely degraded real world is more challenging.  In our proposed method,  the first denoising based on multi-level AdaIN of the manifold of LQ face\cite{adain} has removed most of the degradation patterns such as Gaussian blurring, motion blurring, image noise, and JPEG compaction. Note that Adaptive feature modulation has proven highly effective in a multitude of synthesis tasks, e.g., SPADE \cite{spade2019}, StyleGAN \cite{style2019} \cite{stylegan2_2020}, FaceShifter \cite{faceshifter_2020}, FaceInpainter \cite{faceinpainter}. We find it's more stable for BFR compared with feature concatenation manner\cite{diffbir}.

There is additional degraded information on Stage I, so it is supposed to further remove the noised information for BIR. We apply information bottleneck to deal with this information compression, i.e.,  optimizing a trade-off between information preservation and restoration quality based on relevant information. DiffMAC is based on a latent diffusion model that first manipulates perceptual image compression to the manifold space, then latent diffusion modeling. Manifold representation learning is equivalent to image-level representation learning with the pre-trained VAE. Therefore, we design the information bottleneck (IB) on the manifold level. An effective IB needs to maximize the mutual information (MI) between the hallucinated and real manifold while minimizing MI between the hallucinated and compressed information. Model details of our approach are in Section \ref{section:app}.

As shown in Fig \ref{fig:m3}, naive two denoising with manifold AdaIN (w/o MIB) has lots of artifacts on hair, face, and background, MIB w/o id information compensation fails to preserve the identity of Stage I. Conversely, DID pipeline allows DiffMAC to comprehensively preserve both the structure and detailed textures by taking into account the noise info compression and id info injection. As a result, DiffMAC can effectively synthesize high-fidelity faces for diverse scenes.

\begin{figure*}[htbp]
\begin{center}
   \includegraphics[width=0.92\linewidth]{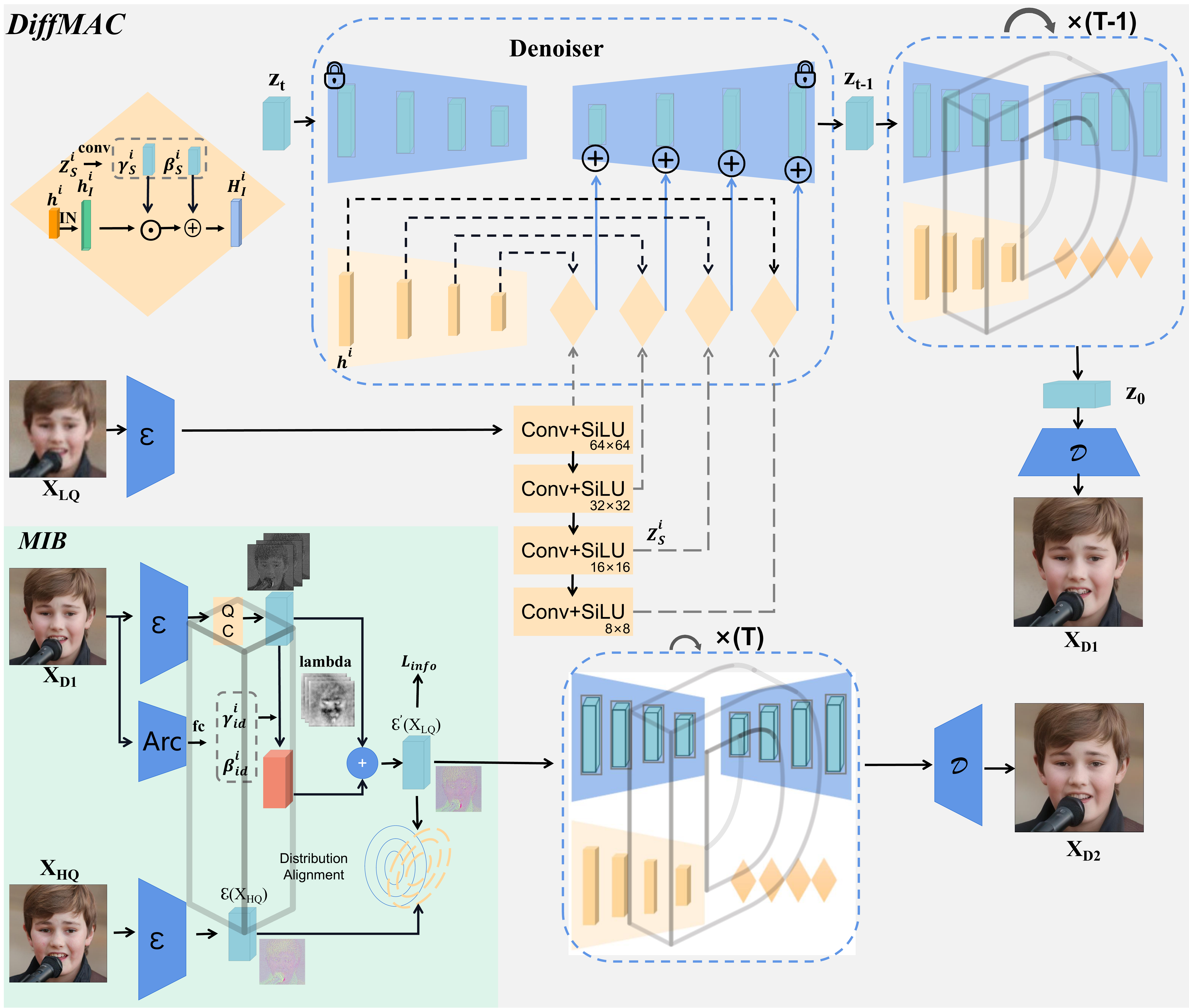}
\end{center}
   \caption{DID framework adopts a two-stage strategy for diffusion finetuning to tackle the challenging HG-BFR task, encompassing one AdaIN-based diffusion and another with MIB. We implement diffusion manifold hallucination correction (DiffMAC) from distorted manifold $\mathcal{E}_{X_{LQ}}$ to the optimized manifold $\mathcal{E}_{X_{D1}}$ for Stage I and $\mathcal{E}^{'}(X_{LQ})$, i.e., $Z$ in Algorithm \ref{alg:alg1}, for Stage II based on information bottleneck. The compressed manifold with identity injection is then used to accurately control feature transformations of the pre-trained stable diffusion model finetuned by Stage I. QC means $quant\_conv$ layer of the encoder of pre-trained VAE. Overall, the DiffMAC framework effectively achieves high-quality BFR results with robustness in diverse domains.}
\label{fig:pipeline}
\end{figure*}
Our paper presents several significant contributions, including:
(1) We propose a high-generalization framework called DID (Diffusion-Information-Diffusion) to address the challenging diffusion manifold hallucination correction (DiffMAC) task. By taking into account modulating diffusion model prior in an AdaIN manner as well as manifold information bottleneck, our approach achieves exceptional BFR results for both synthetic and real-world LQ images with severely degraded scenes. Furthermore, DiffMAC demonstrates remarkable generalization capabilities in heterogeneous domains.
(2) We present an efficient and effective Manifold Information Bottleneck (MIB) module, which provides a trade-off between diffusion manifold maintenance and compression, which disentangles restoration-relevant manifold and degradations. This novel approach improves the model's interpretability and controllability.
(3) We first study the challenging manifold hallucination correction problem for high-generalization BFR task, which is meaningful for task retargeting of foundation model, such as stable diffusion. Compared with state-of-the-art model-based and dictionary-based approaches, DiffMAC demonstrates competitive performance in fidelity and quality for photorealistic face-in-the-wild datasets and HFW (heterogeneous face-in-the-wild) dataset.

\section{Related Work}
\label{gen_inst}

\subsection{Model-based Image Restoration} 
Several remarkable methods that leverage pre-trained StyleGAN \cite{stylegan2_2020} as the face prior have been proposed recently, including GLEAN \cite{bank2021}, GFP-GAN \cite{gfp2021}, PULSE \cite{pulse2020}, GPEN \cite{GPEN}, et al. Another kind of promising approach is based on a stable diffusion model, such as DifFace \cite{difface}, and DiffBIR \cite{diffbir} that contains two stages respectively for degradation removement based on SwinIR and denoising refinement based on diffusion mechanism. Note that our DiffMAC also is on this way.

\subsection{Reference-based Image Restoration} 
Reference-based methods with guided images have been proposed as well, such as WarpNet \cite{warpnet2018}, ASFFNet \cite{exemplar2020}, CIMR-SR\cite{pool2020}, CPGAN \cite{copy2020}, Masa-sr \cite{masa2021}, PSFRGAN \cite{psfr}. These methods explore and incorporate the priors of reference images,  e.g., facial landmarks, identity, semantic parsing, or texture styles, for adaptive feature transformation. Our DiffMAC utilizes a universal prior with Gaussian distribution to modulate the degraded image based on information bottleneck.

\subsection{Bank-based Image Restoration}
These methods first establish high-quality feature dictionaries based on VQGAN \cite{vqgan} or VQVAE \cite{vqvae}, followed by feature matching for image restoration, for example, DFDNet \cite{dfdnet2020}, Codebook Lookup Transformer (CoLT) \cite{codeformer},  VQFR \cite{vqfr}. These methods rely so highly on the prior bank that having low generalization across heterogeneous domains. In contrast, our DiffMAC operates restoration without any prior bank.
\subsection{Information Compression}
\cite{ib} first proposes an information bottleneck (IB)  that takes a trade-off between information compression and robust representation ability for specific tasks. \cite{vib} leverage the reparameterization trick \cite{trick} with variational approximation to efficiently train the IB neural layer. \cite{iba} restricts the attribution information using adaptive IB for effective disentanglement of classification-relative and irrelative information. InfoSwap \cite{infoswap} views face swapping in the plane of information compression to generate identity-discriminative faces. Furthermore, \cite{infopaint} uses highly compressed representation that maintains the semantic feature while ignoring the noisy details for better image inpainting. We will introduce our manifold information bottleneck in Section \ref{section:mib}.

\section{Approach}
\label{section:app}

In this section, we provide a detailed introduction to our proposed DiffMAC method, including the overall DID framework in Sec. \ref{section:did}, along with the manifold information bottleneck (MIB) module in Sec. \ref{section:mib}. As shown in Figure \ref{fig:pipeline}, given a target face $X_{HQ}$, we obtain the degraded image $X_{LQ}$ using sequential degradation methods such as blurring, downsampling, noise injection, and JPEG compression. We implement manifold hallucination correction making distorted manifold $\mathcal{E}_{LQ}$ close to the real manifold $\mathcal{E}_{HQ}$ based on information bottleneck. The MIB module produces disentangled attention spatial maps for the original hallucinated manifold and injected information. Subsequently, we obtain multi-scale spatial features $z_{S}=\{z_{S}^{1}, z_{S}^{2}, ..., z_{S}^{n}\}$ from manifold, and impose them on corresponding UNet layer to modulate the pre-trained stable diffusion model.

\subsection{DiffMAC}
\label{section:did}
Latent diffusion model \cite{latentdiff} conducts diffusion process on the compressed latent, i.e., manifold of the image distribution. This efficient manner is also leveraged in ControlNet \cite{controlnet} and DiffBIR \cite{diffbir}. The distribution constraint of the latent diffusion model is:
\begin{equation}
\mathcal{L}_{ldm}=\mathbb{E}_{z,c,t,\epsilon}[\Vert\epsilon-\epsilon_{\theta}(z_{t}=\sqrt{\overline{\alpha}_{t}}z+\sqrt{1-\overline{\alpha}_{t}}\epsilon, c, t)\Vert_{2}^{2}],
\end{equation},
where $z$ denotes the manifold obtained via encoder of VAE, i.e., $z=\mathcal{E}(I_{HQ})$.
$\epsilon \sim \mathcal{N}(0,\mathbb{I})$ with variance $\beta_{t}=1-\alpha_{t}\in(0,1)$ that is used to generate noisy manifold.

In our proposed DiffMAC approach, we finetune the stable diffusion model with the multi-level spatial feature alignment based on $\mathcal{E}_{LQ}$. The input feature $h^{i}$ in each stable diffusion (SD) encoder block is modulated through affine transform parameters derived from $z_{S}$ in $i$th level that can be expressed as:

\begin{equation}
h^{i}_{I}=\frac{h^{i}-\mu^{i}_{I}}{\sigma ^{i}_{I}},
\end{equation}
\begin{equation}
H^{i}_{I}=\gamma _{S_{I}}^{i}\odot h^{i}_{I} +\beta _{S_{I}}^{i},
\end{equation}
where  $h^{i} \in \mathbb{R}^{C_{h}^{i}\times H^{i} \times W^{i}}$, $\mu^{i}_{I}$ and $\sigma^{i}_{I}$ are the means and standard deviations of $h^{i}$, and they are utilized to conduct the instance normalization. $\gamma _{S_{I}}^{i}$ and $\beta _{S_{I}}^{i}\in \mathbb{R}^{C_{h}^{i}\times H^{i}\times W^{i}}$ are obtained from $z_{S}^{i}$ using respective convolutional layer. As shown in Figure \ref{fig:pipeline}, spatial features from LQ faces are obtained using a ConvLayer and SiLU activation. We employ zero convolution with both weight and bias initialized as zeros \cite{controlnet}. The aligned spatial feature of SD encoder block $H^{i}$ is formulated as
\begin{equation}
H^{i}=H^{i}_{I} + h^{i}.
\label{eq3}
\end{equation}

Once the optimization of the latent diffusion model is finished, the aligned manifold is calculated as follows:
\begin{equation}
\Tilde{z}_{0}=\frac{z_{t}}{\sqrt{\overline{\alpha}_{t}}}-\frac{\sqrt{1-\overline{\alpha}_{t}}\epsilon_{\theta}(z_{t},c,t)}{\sqrt{\overline{\alpha}_{t}}}.
\label{eq3}
\end{equation}

Algorithm pseudo code is described in Algorithm \ref{alg:alg1}. The image-level output $X_{D1} = \mathcal{D}(z^{1}_{0})$ with decoder of pretrained VAE.

We propose to use manifold information compression on $\mathcal{E}_{X_{D1}}$ to extract the valuable manifold that produces $z_{S}^{i}$ for the AdaIN transformation of Stage II.
\begin{figure*}[htbp]
\begin{center}
   \includegraphics[width=0.94\linewidth]{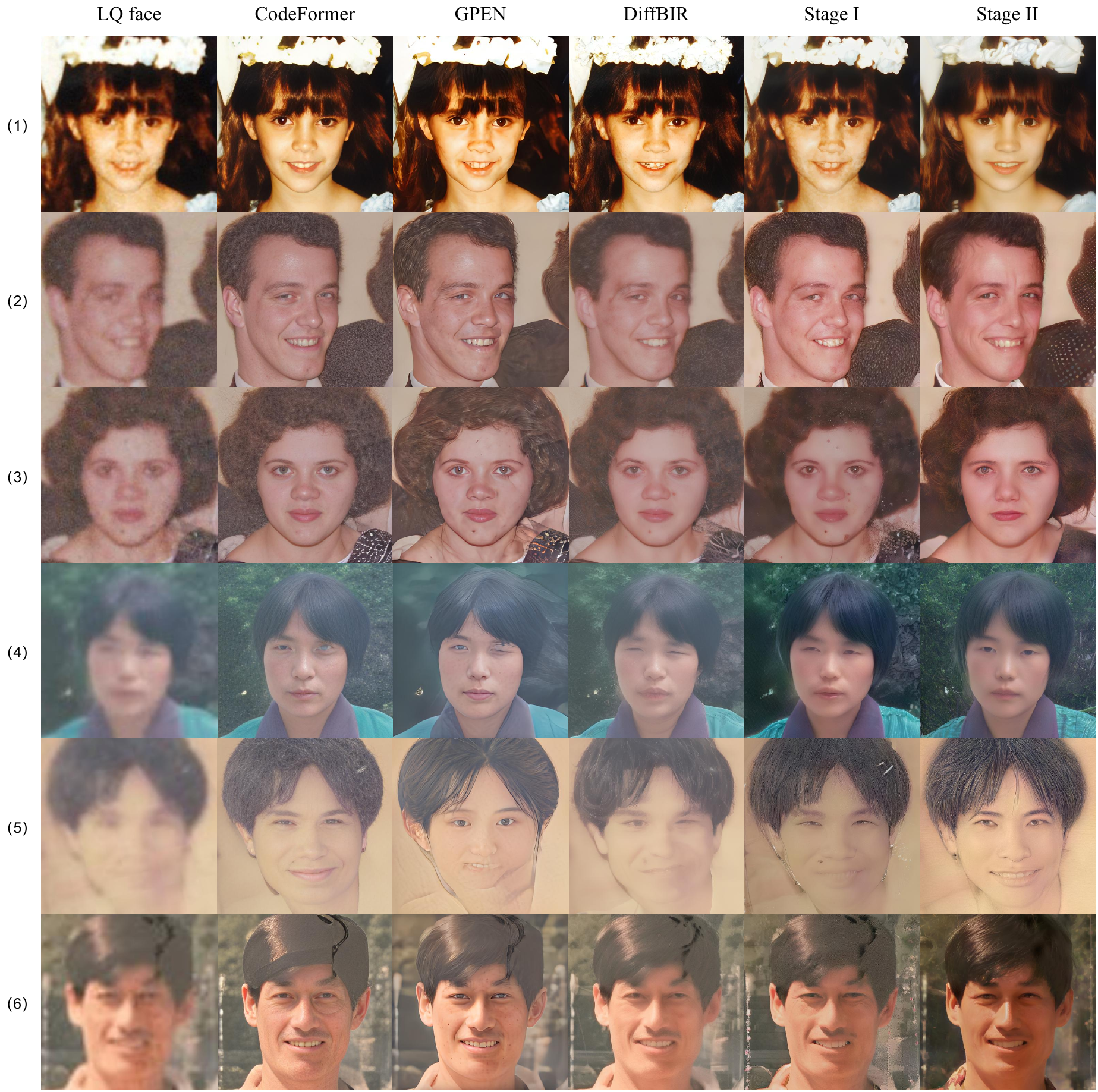}
\end{center}
   \caption{MIB plays an important role in severely degraded scenarios. Stage II demonstrates more natural and cleaner than Stage I beleaguered by facial noises and ambiguous textures. Moreover, CodeFormer \cite{codeformer} has messy hair (row $2\&3\&5$), and incompatible eyes (row 4$\&$6). GPEN \cite{GPEN} has more obvious artifacts, and DiffBIR \cite{diffbir} indicates more blurry imaging, in contrast to Stage II of DiffMAC. }
\label{fig:stage2} 
\end{figure*}

\subsection{Manifold Information Bottleneck}
\label{section:mib}
MIB is designed as a powerful function that improves relevant information between the hallucinated and real manifold, while maximally compressing the restoration-irrelevant perceptual information caused by diverse degradation patterns in it. MIB can be formulated as follows:
\begin{equation}
\label{equ:mib}
\mathop{\min}_{Z} \beta\mathbb{I}(\mathcal{E}_{X_{D1}};Z)-\mathbb{I}(Z;\mathcal{E}_{X_{HQ}})
\end{equation},
where $\mathbb{I}$ means the mutual information function, $Z$ is the optimal representation based on $\mathcal{E}_{X_{D1}}$.

To extract the useful manifold and stay away from the noised manifold, we design an information filter $\lambda$. Furthermore, to address uncontrollable identity distortion, we incorporate compressed manifold and identity modulated manifold $\epsilon_{id}$ as the optimized manifold. Given $R = \mathcal{E}_{X_{D1}}$, $\epsilon_{id}$ is formulated as:

\begin{equation}
\mathcal{M}_{I}=\frac{R-\mu_{\mathcal{M}}}{\sigma_{\mathcal{M}}},
\end{equation}
\begin{equation}
\epsilon_{id}=\gamma _{id}\odot \mathcal{M}_{I} +\beta _{id},
\end{equation}
where  $R \in \mathbb{R}^{C^{M}\times H^{M}\times W^{M}}$, $\mu_{\mathcal{M}}$ and $\sigma_{\mathcal{M}}$ represent the means and standard deviations of $R$, and they are utilized to conduct AdaIN transformation. $\gamma _{id}$ and $\beta _{id}\in \mathbb{R}^{C^{M}}$ are obtained from identity embedding from the pre-trained face recognition model using respective fully-connected layers. $\epsilon_{id}$ is effective to prevent the optimized manifold not deviating too far from its original identity.

\begin{algorithm}[tbp]
\caption{DID algorithm concerning DiffMAC}\label{alg:alg1}
\SetKwInOut{Input}{Input}\SetKwInOut{Output}{Output}
\SetKwInOut{Require}{Require}
\Input{$X_{LQ}$, $X_{HQ}$, diffusion step $T$, text prompt $c$ (set to " ")}
\Require{Latent diffusion model $\epsilon_{\theta}$ (Stable diffusion v2.1), VAE's encoder $\mathcal{E}$, $quant\_conv$ $QC$, and decoder $\mathcal{D}$, pretrained ArcFace recognition model $\mathbb{F}_{ID}$}
\Require{Pre-calculated mean $\mu_{QC}$ and std. $\sigma_{QC}$ values}
\Output{Restored face $X_{D2}$}
Sample $\epsilon_{1}$ from $\mathcal{N}(0,\mathbb{I})$\;
\For(\hfill{$\triangleright \ $1st Diff}){$t\leftarrow 1$ \KwTo $T$} 
{
$\mathcal{L}_{ldm}^{1}\leftarrow 0$\;
$R\leftarrow \mathcal{E}_{X_{LQ}}$, $z\leftarrow \mathcal{E}_{X_{HQ}}$\;
$\mathcal{R}\leftarrow $ DiagonalGaussianDistribution($R$)\;
\If{$l$ is ControlLayer}
{
AdaIN based on multi-level features of $\mathcal{R}$;}
$z_{t}\leftarrow\sqrt{\overline{\alpha}_{t}}z+\sqrt{1-\overline{\alpha}_{t}}\epsilon_{1}^{\theta}(z_{t},c,t, \mathcal{R})$\;
$\mathcal{L}_{ldm}^{1}\leftarrow \mathbb{E}[\epsilon_{1}, \epsilon_{1}^{\theta}(z_{t},c,t, \mathcal{R})]$\;}

Sample $\Tilde{z}_{0}^{1}\leftarrow \frac{z_{t}}{\sqrt{\overline{\alpha}_{t}}}-\frac{\sqrt{1-\overline{\alpha}_{t}}\epsilon_{1}^{\theta}(z_{t},c,t, \mathcal{R})}{\sqrt{\overline{\alpha}_{t}}}$\;
$X_{D1}\leftarrow \mathcal{D}(z_{0}^{1})$\;
Sample $\epsilon_{2}$ from $\mathcal{N}(0,\mathbb{I})$\;
\For{$t\leftarrow 1$ \KwTo $T$} 
{
$\mathcal{L}_{info}\leftarrow 0$, $\mathcal{L}_{rec}\leftarrow 0$, $\mathcal{L}_{ldm}^{2}\leftarrow 0$\;%
$R\leftarrow \mathcal{E}(X_{D1})$, $z\leftarrow \mathcal{E}_{X_{HQ}}$\; 
$\mathbb{F}_{R}\leftarrow \frac{R-\mu_{QC}}{max(\sigma_{QC},o)}$\;
$\mu_{ID}\leftarrow FC(\mathbb{F}_{ID}(X_{D1}))$, $\sigma_{ID}\leftarrow FC(\mathbb{F}_{ID}(X_{D1}))$\;
$\mathbb{F}_{N}\leftarrow \mathbb{F}_{R}*\sigma_{ID}+\mu_{ID}$\;
$\lambda \leftarrow Sigmoid(Conv(\mathbb{F}_{R}))$\;
$Z\leftarrow \lambda*R+(1-\lambda)*\mathbb{F}_{N}$;\hfill{$\triangleright \ $MIB}\ 

$\mathcal{L}_{info}\leftarrow \mathbb{I}[Z,R]$\;
$Z\leftarrow $ DiagonalGaussianDistribution($Z$)\;
$z\leftarrow $ DiagonalGaussianDistribution($z$)\;
$\mathcal{L}_{rec}\leftarrow \mathbb{I}[Z,z]$\;
\If{$l$ is ControlLayer}
{
AdaIN based on multi-level features of $Z$;}
$z_{t}\leftarrow\sqrt{\overline{\alpha}_{t}}z+\sqrt{1-\overline{\alpha}_{t}}\epsilon_{2}^{\theta}(z_{t},c,t, Z)$;\hfill{$\triangleright \ $2nd Diff}\

$\mathcal{L}_{ldm}^{2}\leftarrow \mathbb{E}[\epsilon_{2}, \epsilon_{2}^{\theta}(z_{t},c,t, Z)]$\;}

Sample $\Tilde{z}_{0}^{2}\leftarrow \frac{z_{t}}{\sqrt{\overline{\alpha}_{t}}}-\frac{\sqrt{1-\overline{\alpha}_{t}}\epsilon_{2}^{\theta}(z_{t},c,t, Z)}{\sqrt{\overline{\alpha}_{t}}}$\;
$X_{D2}\leftarrow \mathcal{D}(z_{0}^{2})$.\

\end{algorithm}

Then, the optimized manifold can be formulated as follows:
\begin{equation}
Z = \lambda R+(1-\lambda)\epsilon_{id}
\label{equ:fuse}
\end{equation},
where R is from $\mathcal{N}(\mu_{\mathcal{M}},\sigma_{\mathcal{M}}^{2})$ since the manifold $\mathcal{M}$ extraction module of latent diffusion model has been trained and fixed. Note that if $\lambda$ is 0, the whole manifold, i.e., $\mathcal{E}_{X_{D1}}$ will be replaced by id information, which results in uncontrollable restoration for the 2nd Diffusion finetuning. If $\lambda$ is 1, adaptive id info will not be integrated to conduct high-fidelity manifold hallucination correction. Qualitative analyses are illustrated in Fig \ref{fig:ablation} and Fig \ref{fig:12}.

The supervision for the manifold representations is formulated as:
\begin{equation}
\mathcal{L}_{rec} = -\mathbb{I}(Z;\mathcal{E}_{X_{HQ}})= ||Z-\mathcal{E}_{X_{HQ}}||_{2}^{2}
\end{equation},
where $Z\approx\mathcal{E}_{HQ}$ whose distribution is $\mathcal{N}(\mu_{\mathcal{M}},\sigma_{\mathcal{M}}^{2})$, so we define $Z\sim \mathcal{N}(\mu_{\mathcal{M}},\sigma_{\mathcal{M}}^{2})$ as well.

Given the distribution of $p(Z|R) \sim \mathcal{N}[\lambda R+(1-\lambda)\mu_{\mathcal{M}},(1-\lambda)^{2}\sigma_{\mathcal{M}}^{2}]$, then the information compression metric is formulated as:
\begin{equation}
\begin{aligned}
\mathcal{L}_{info} &= \mathbb{I}(Z;R) \\
                   &= KL[p(Z|R)\Vert p(Z)]\\
                   &= -log(1-\lambda)^{2}+\frac{1}{2}[(1-\lambda)^{2}+(\lambda\frac{R-\mu_{R}}{\sigma_{R}})^{2}-1],
\end{aligned}
\end{equation}
based on which KL--divergence controls the manifold distance. For example, $KL$==0 $\rightarrow p(Z|R)\sim\mathcal{N}(\mu_{\mathcal{M}},\sigma_{\mathcal{M}}^{2})\rightarrow\lambda==0$, which means the percent of information compression of manifold is 100$\%$. We finetune DiffMAC for further controllable face syntheses according to the optimized manifold.

Finally, the total loss of DiffMAC is formulated as:
\begin{equation}
\mathcal{L}_{\emph{Stage I}}= \mathcal{L}_{ldm}^{1},
\label{con:20}
\end{equation}
\begin{equation}
\mathcal{L}_{\emph{Stage II}}= \mathcal{L}_{ldm}^{2}  +  \lambda_{info}\mathcal{L}_{info}+  \lambda_{rec}\mathcal{L}_{rec}.
\label{con:20}
\end{equation}

\section{Experiment}
\label{others}
\subsection{Experimental Protocol}
\subsubsection{Training Protocol} 
\begin{figure*}[htbp]
\begin{center}
   \includegraphics[width=1\linewidth]{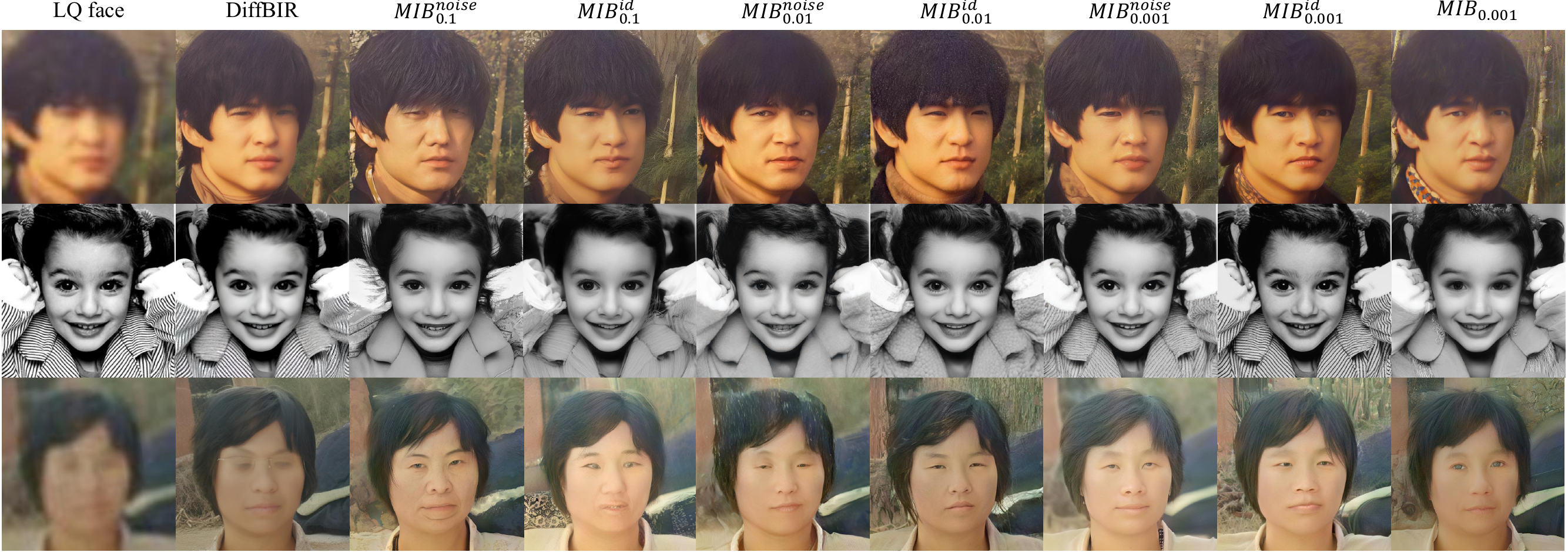}
\end{center}
   \caption{Ablation study concerning different $\beta$ (i.e., 0.1, 0.01, and 0.001) in Equ. \ref{equ:mib} and the additional injected information (i.e., identity or noise) after information compression. $MIB_{0.001}^{id}$ successfully carries on pleasant and high-fidelity restoration with identity preservation. Nevertheless, $MIB_{0.001}$ without information compensation results in undesired structure and texture distortion (row 2). Note that these ablated models are all two-stage.}
\label{fig:ablation} 
\end{figure*}

\begin{figure*}[htbp]
\begin{center}
   \includegraphics[width=0.5\linewidth]{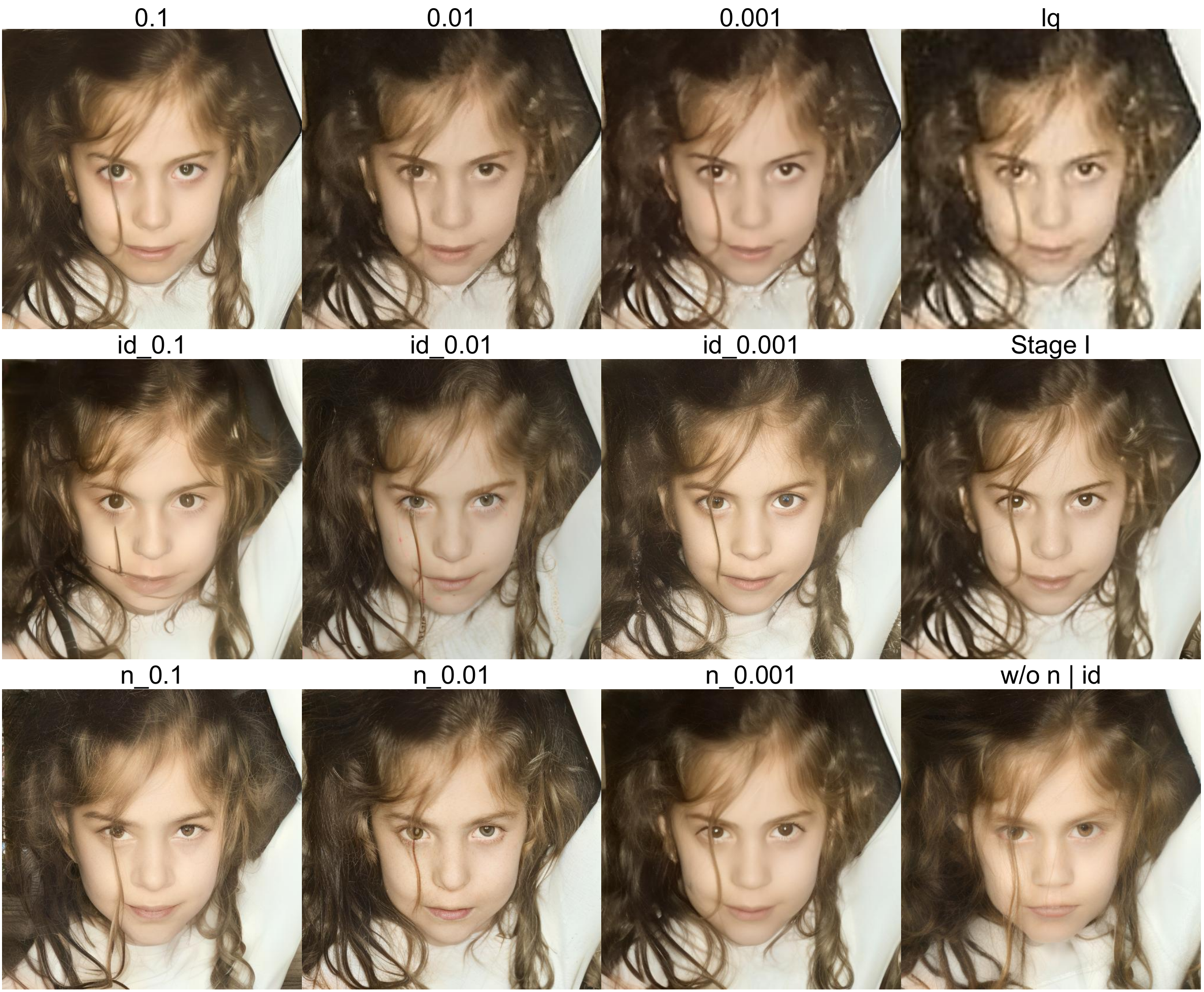}\includegraphics[width=0.5\linewidth]{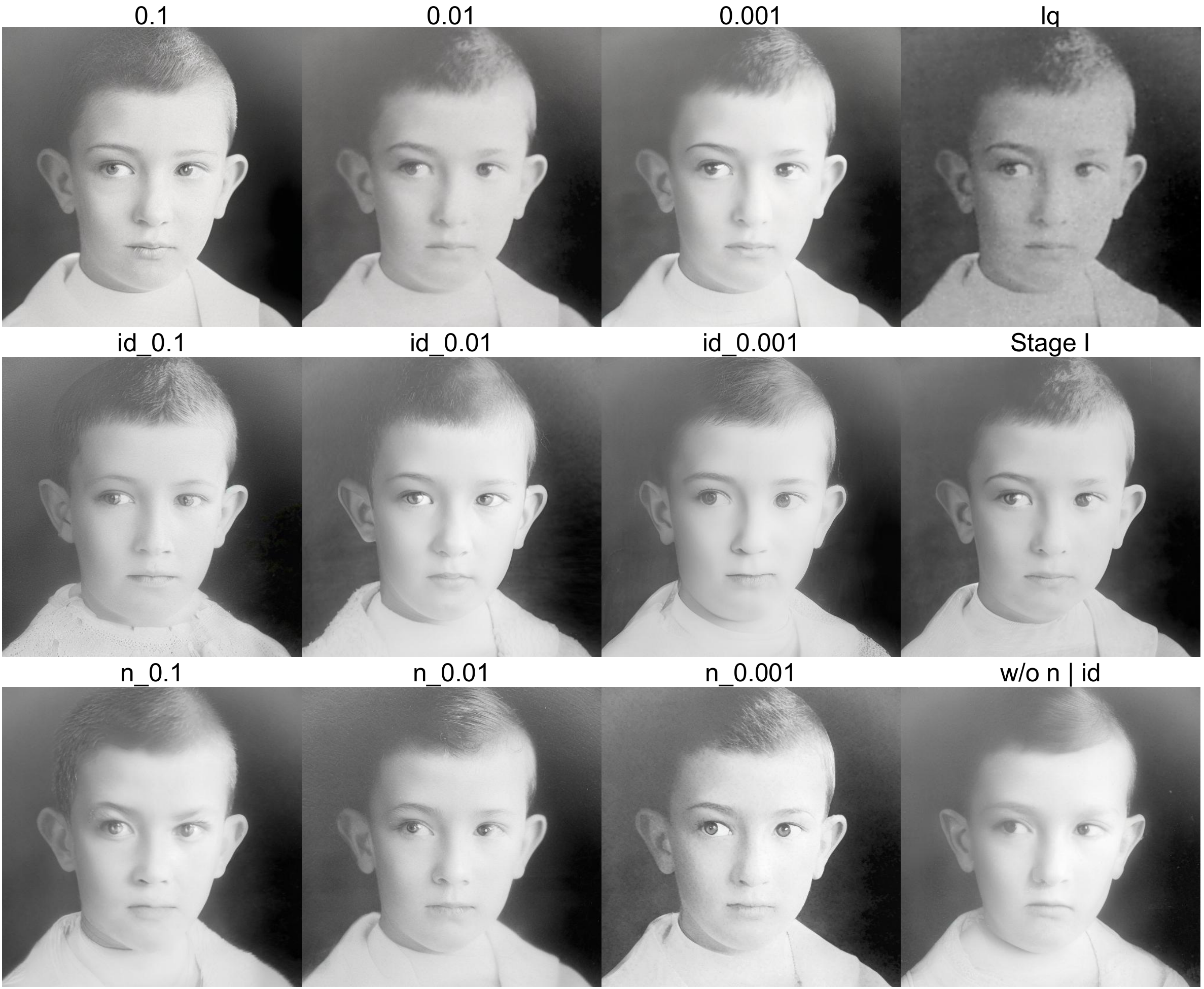}
\end{center}
   \caption{Ablation study concerning 1) MIB w/o Stage I, w/ noise injection denoted by 0.1, 0.01, 0.001 (row 1), 2) MIB w/ id injection based on Stage I (row 2), 3) MIB w/ noise injection based on Stage I (row 3), as well as 4) w/o noise or id injection based on Stage I. w/o Stage I is not capable of generating sharp details, and not friendly to selection of compression degree ($\beta$) of MIB, due to diverse degradation levels of LQ images. w/ id injection with low compression degree (i.e., id$\_$0.001) is more competent to synthesize high-quality structure and texture details, compared with id$\_$0.1. }
\label{fig:12} 
\end{figure*}

We train our DiffMAC on the FFHQ \cite{style2019} dataset with $512\times512$ size.  We train Stage I and Stage II for 200k and 10k iterations respectively with one NVIDIA RTX 4090 GPU. The training batch size is set to 2. We utilize Stable Diffusion 2.1 as the generative prior. During training, we employ AdamW \cite{adamw} with $10^{-4}$ learning rate. We adopt spaced DDPM sampling \cite{ddpm} with 50 timesteps for denoising inference. To better compress the model parameters, the two stages share the same VAE for manifold encoding and decoding. We set $\lambda_{rec}=1, \lambda_{info}=0.001$. Larger $\lambda_{info}$ leads to more information missing of $X_{D1}$. More ablation experiments are demonstrated in Fig \ref{fig:ablation} and Fig \ref{fig:12}. Moreover, the manifold feature after QC (quant$\_$conv layer) has 8 channels with same setting as $\gamma _{id}$, $\beta _{id}$ and information filter $\lambda$. Note that minimal std $o$ in Algorithm \ref{alg:alg1} is set to 0.01.
\begin{figure*}[htbp]
\begin{center}
   \includegraphics[width=0.85\linewidth]{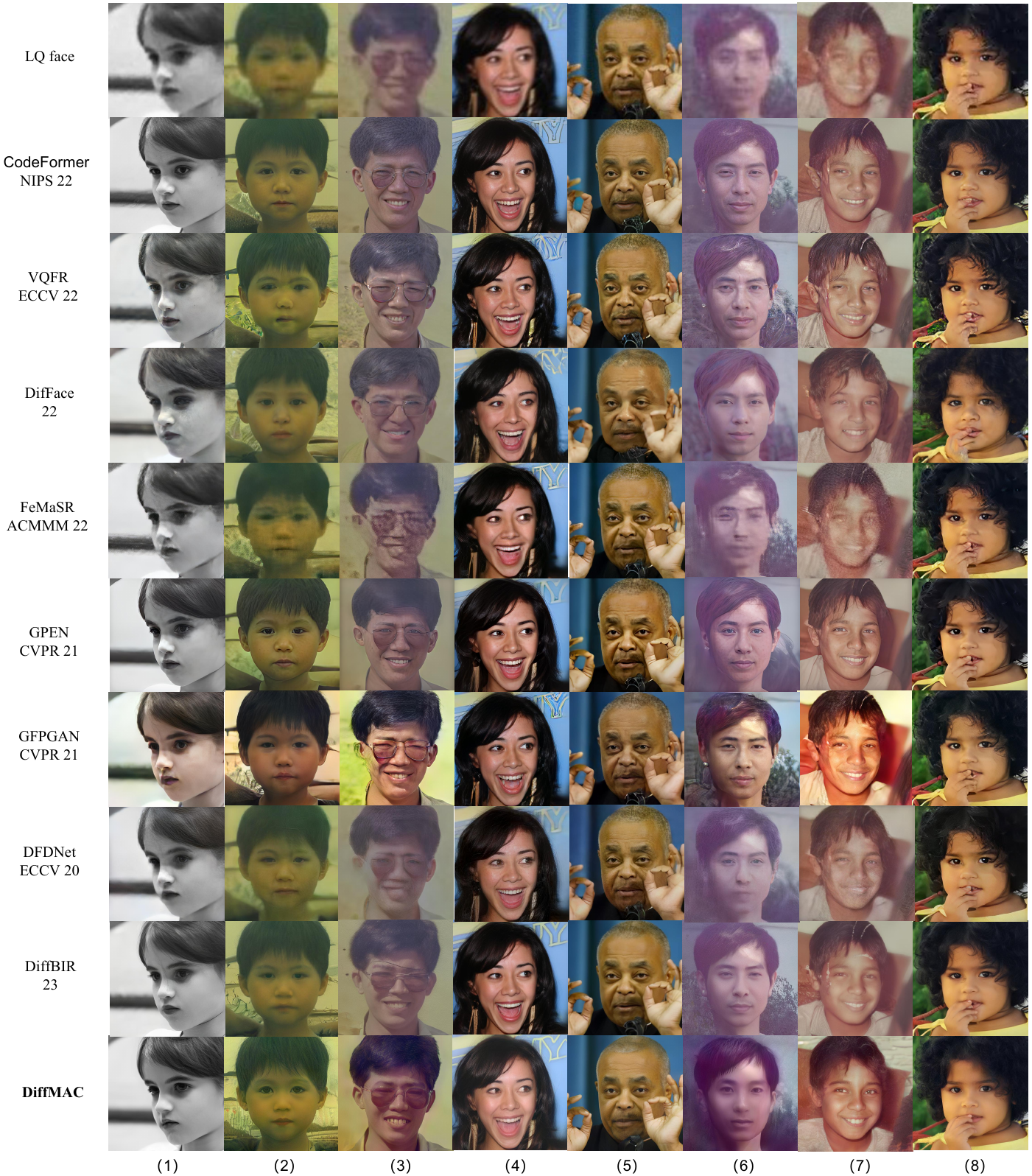}
\end{center}
   \caption{Qualitative comparison with state-of-the-art methods (CodeFormer \cite{codeformer}, VQFR \cite{vqfr}, DifFace \cite{difface}, FeMaSR \cite{femasr}, GPEN \cite{GPEN}, GFPGAN \cite{gfp2021}, DFDNet \cite{dfdnet2020} and DiffBIR \cite{diffbir}) in severely degraded photorealistic scenes. DiffMAC is capable of cleaner background$\&$face (col 1$\&$7), imaginative background complement (col 2), higher-fidelity sunglass (col 3), natural extreme expression (col 4), more realistic hairstyle (col 6), and more stable reconstruction with facial occlusion (col 5$\&$8). Zoom in for better observation.}
\label{fig:photo}
\end{figure*}

\begin{figure*}[htbp]
\begin{center}
   \includegraphics[width=0.84\linewidth]{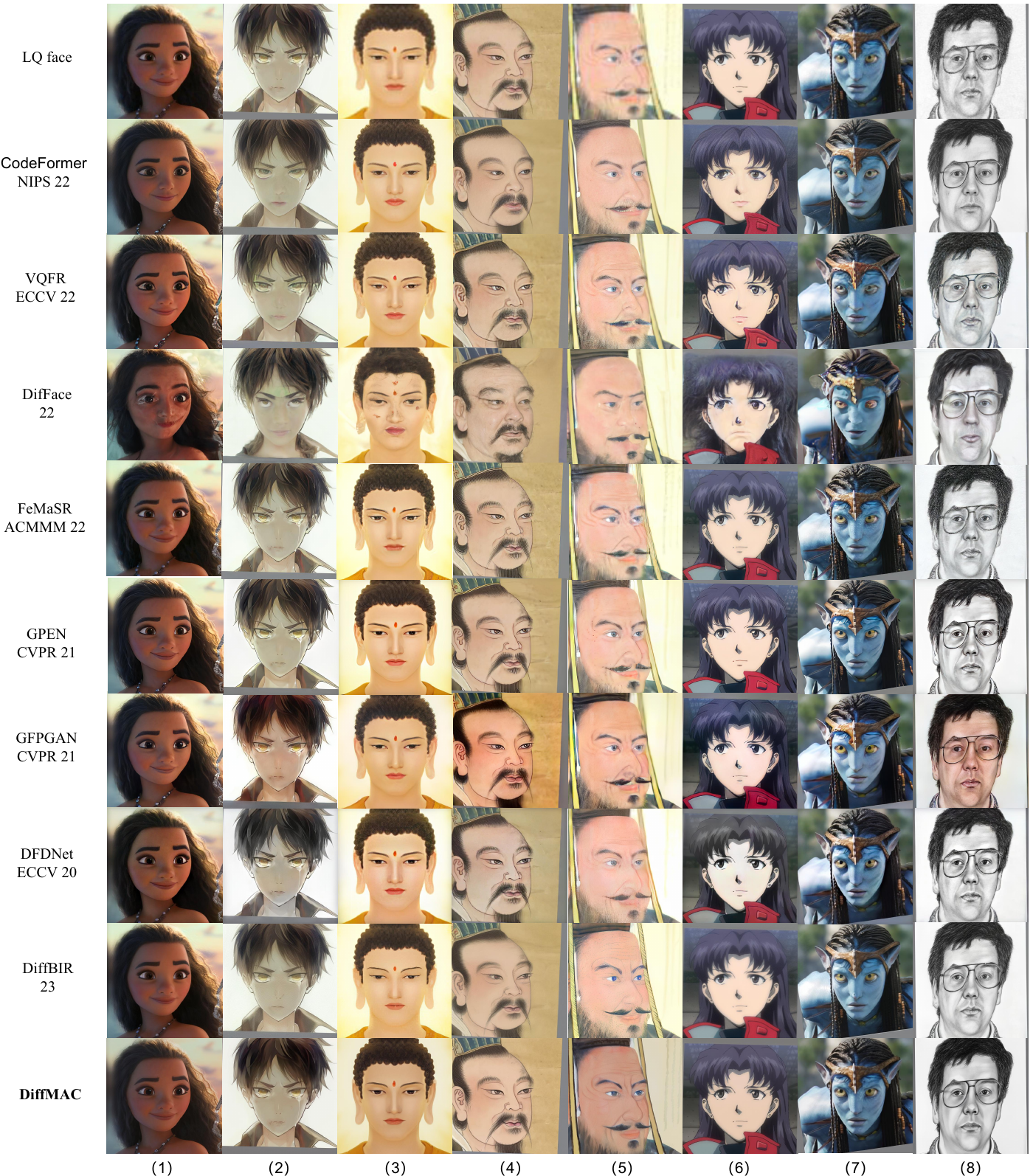}
\end{center}
   \caption{Qualitative comparison with state-of-the-art methods (CodeFormer \cite{codeformer}, VQFR \cite{vqfr}, DifFace \cite{difface}, FeMaSR \cite{femasr}, GPEN \cite{GPEN}, GFPGAN \cite{gfp2021}, DFDNet \cite{dfdnet2020} and DiffBIR \cite{diffbir}) in out-of-distribution degraded heterogeneous scenes. \cite{codeformer,vqfr,difface} synthesize human-like face with human eyes or mouth (col $2\&5\&6$), \cite{femasr} fails to fight against severe and complicated degradations (col 3$\&5\&8$), \cite{GPEN} discards lots of hair details and generates overly sharp lines, \cite{gfp2021, dfdnet2020} have severe color deviation, \cite{diffbir} easily produce line, block or dot artifacts (col $2\&3\&5$). DiffMAC is more stable for heterogeneous BIR. More analyses are illustrated in Fig \ref{fig:heterm}.}
\label{fig:heter}
\end{figure*}

we apply the degradation model adopted in \cite{codeformer}\cite{diffbir} to synthesize the training data with a similar distribution to the real degraded images. The degradation pipeline is as follows:  
\begin{equation}
X_{LQ}=\{[(X_{HQ}\otimes \mathbf{k}_{\mathcal{G}})_{\downarrow r}+\mathbf{n}_{\sigma}]_{JPEG_{q}}\}_{\uparrow r},
\end{equation}
where $\mathbf{k}_{\mathcal{G}}$ denotes Gaussian blur kernel with $\mathcal{G} \in \{1:15\}$. Moreover, the down-sampling scale r, Gaussian noise $\mathbf{n}_{\sigma}$, and JPEG compression quality $q$ are in the range of $\{0.8 : 8\}$, $\{0 : 20\}$, and $\{60 : 100\}$, respectively. These degradation operations are randomly integrated into the training stage.

\subsubsection{Test Datasets} We evaluate the performance of our DiffMAC framework on the synthetic dataset CelebA-HQ \cite{celeb2017}, real-world datasets, i.e., LFW \cite{lfw}, CelebChild \cite{gfp2021}, WebPhoto \cite{gfp2021} and our collected HFW (heterogeneous face in the wild) dataset.

\textbf{CelebA-Test} is a synthetic dataset with 3,000 CelebA-HQ images from its test set \cite{celeb2017}. We directly utilize the degraded images of VQFR \cite{vqfr} for fair comparison.

\textbf{LFW-Test} \cite{lfw} contains 1711 LQ images in the wild where the first image of each person in the LFW dataset is selected.

\textbf{CelebChild\cite{gfp2021}} consists of 180 real-world child pictures in color or grayscale space.

\textbf{WebPhoto\cite{gfp2021}} consists of 407 LQ faces collected on the Internet. Most of them are from Asian faces.

\textbf{HFW} contains 1200 collected heterogeneous faces, e.g., sculpture portrait, Buddha, ancient portrait, oil painting, Beijing opera actors, exaggerated drawings, 2D/3D anime characters, Avatars, NIR, sketch and night-shot portrait.


\subsubsection{Evaluation Metric}

We compare the performance of our method with state-of-the-art. The metrics for CelebA-Test include image quality analyzers (patch-wise perceptual LPIPS \cite{lpips},  FID \cite{FID}, luminance-level NIQE \cite{NIQE}, transformer-based MUSIQ \cite{musiq}), identity consistency measurements (the embedding similarity of Arcface \cite{arcface2019}, landmark localization error (LLE)), as well as classical perception measurements (PSNR and SSIM). A quantitative comparison of the synthetic dataset is shown in Tab \ref{tab:celeb}.

NIQE \cite{NIQE} and MUSIQ \cite{musiq} are non-reference blind image quality assessment methods. Quantitative comparisons of real-world and heterogeneous datasets are shown in Tab \ref{tab:real}. Note that quantitative scores are only reference analyses, especially FID of HFW, e.g., DifFace severely destroys the original heterogeneous content of the original face (Fig \ref{fig:heter}).

\subsubsection{Comparison methods} We evaluate the performance of our DiffMAC framework with 8 recent state-of-the-art methods, i.e., dictionary-based methods (CodeFormer \cite{codeformer}, VQFR \cite{vqfr}, FeMaSR \cite{femasr}, DFDNet \cite{dfdnet2020}), StyleGAN-based methods (GPEN \cite{GPEN}, GFPGAN \cite{gfp2021}),  and diffusion-based methods (DifFace \cite{difface}, DiffBIR \cite{diffbir}). Here we briefly summarise their advantages and limitations.

\begin{table}[htbp]
\centering
\huge
\caption{Quantitative comparison with state-of-the-art methods (CodeFormer \cite{codeformer}, VQFR \cite{vqfr}, DifFace \cite{difface}, FeMaSR \cite{femasr}, GPEN \cite{GPEN}, GFPGAN \cite{gfp2021}, DFDNet \cite{dfdnet2020} and DiffBIR \cite{diffbir}) on CelebA-Test.}
\scalebox{0.35}{
\begin{tabular}{c|cc|cc|cc|cc}
\hline 
 Methods & LPIPS$\downarrow$ & FID$\downarrow$ & NIQE $\downarrow$ & MUSIQ$\uparrow$ & Arc$\uparrow$ & LMD$\downarrow$ & PSNR$\uparrow$ & SSIM$\uparrow$\tabularnewline
\hline 
\hline
Input &0.4866 	&143.98 	&13.440	&27.71	&0.65	&3.76 	&25.35 	&0.6848\tabularnewline

\hline 
CodeFormer &	\textcolor{red}{0.3432}	&52.45	&4.648	&\textcolor{red}{71.82}	&0.79	&2.47	&25.14	&0.6698	\tabularnewline

VQFR &\textcolor{blue}{0.3515}  	&\textcolor{blue}{41.28}	&\textcolor{red}{3.693}	&\textcolor{blue}{71.51}	&0.80	&2.43  	&24.14	&0.6360 \tabularnewline

DifFace &	0.3910	&\textcolor{red}{40.08}	&4.504	&67.69	&0.64	&3.26	&24.15	&0.6668\tabularnewline

FeMaSR&	0.3899	&122.49	&4.952	&68.90	&0.77	&2.82	&25.28	&0.6507\tabularnewline

GPEN &	0.3667	&45.42	&5.729	&71.40	&0.81	&2.53	&24.60	&\textcolor{blue}{0.6790}\tabularnewline

GFPGAN &	0.3646 	&42.62 	&\textcolor{blue}{4.077}	&70.80	&0.81	&2.41 	&25.08 	&0.6777\tabularnewline

DFDNet &	0.4341  	&59.08	&4.341	&70.56	&0.75	&3.31 	&23.68 	&0.6622\tabularnewline
\hline
DiffBIR &	0.3783	&48.05	&5.458	&69.60	&\textcolor{blue}{0.81}	&\textcolor{blue}{2.25}	&\textcolor{blue}{25.58}	&\textcolor{red}{0.6793}\tabularnewline

DiffMAC &	0.4035	&47.15	&5.812	&69.26	&\textcolor{red}{0.82}	&\textcolor{red}{2.21}	&\textcolor{red}{25.59}	&0.6734\tabularnewline
\hline
GT &	0	&43.43	&4.372	&70.38	&1	&0	&$\infty$	&1\tabularnewline

\hline 
\end{tabular}}

\label{tab:celeb}
\end{table}
\begin{table*}[htbp]
\centering
\caption{Quantitative evaluation compared with state-of-the-art methods (CodeFormer \cite{codeformer}, VQFR \cite{vqfr}, DifFace \cite{difface}, FeMaSR \cite{femasr}, GPEN \cite{GPEN}, GFPGAN \cite{gfp2021}, DFDNet \cite{dfdnet2020} and DiffBIR \cite{diffbir}) on LFW-Test, CelebChild, WebPhoto and HFW.}
\scalebox{0.83}{
\begin{tabular}{c|ccc|ccc|ccc|ccc}
\hline 
\makecell[c]{Datasets}  & \multicolumn{3}{c|}{LFW-Test} & \multicolumn{3}{c|}{CelebChild} & \multicolumn{3}{c|}{WebPhoto} &\multicolumn{3}{c}{HFW}\tabularnewline
\hline 
\makecell[c]{Methods} & \makecell[c]{FID$\downarrow$} & \makecell[c]{NIQE$\downarrow$} & \makecell[c]{MUSIQ$\uparrow$} & \makecell[c]{FID$\downarrow$} & \makecell[c]{NIQE$\downarrow$} & \makecell[c]{MUSIQ$\uparrow$}& \makecell[c]{FID$\downarrow$} & \makecell[c]{NIQE$\downarrow$} & \makecell[c]{MUSIQ$\uparrow$}& \makecell[c]{FID} & \makecell[c]{NIQE$\downarrow$} & \makecell[c]{MUSIQ$\uparrow$}\tabularnewline
\hline 
\hline
Input &137.56 	&11.214	&25.05 	&144.42 	&9.170	&47.85	&170.11 	&12.755	&19.24	&108.02	&8.30	&51.54\tabularnewline

\hline 
CodeFormer &52.02	&4.483	&\textcolor{blue}{71.43}	&116.24	&4.981	&70.83	&\textcolor{blue}{78.87}	&4.709	&70.51	&\textcolor{blue}{76.34}	&4.931	&70.54\tabularnewline

VQFR &50.64 	&\textcolor{red}{3.589}	&71.39	&\textcolor{red}{105.18} 	&\textcolor{red}{3.936}	& \textcolor{blue}{70.92}	&\textcolor{red}{75.38} 	&\textcolor{red}{3.607}	&\textcolor{blue}{71.35}	&78.38	&\textcolor{red}{4.094}	&69.72\tabularnewline

DifFace &	45.18	&4.285	&69.65	&\textcolor{blue}{108.71}	&4.570	&68.04	&87.05	&4.675	&67.87	&\textcolor{red}{63.19}	&4.777	&65.47\tabularnewline

FeMaSR&118.21	&4.802	&68.97	&136.13	&5.832	&66.54	&149.98	&5.236	&57.09	&219.69	&6.201	&\textcolor{blue}{71.28}\tabularnewline

GPEN &	57.58 	&5.644	&\textcolor{red}{73.59}	&121.11	&5.883	&\textcolor{red}{71.42}	&81.77	&6.170	&\textcolor{red}{73.41}	&104.24	&6.182	&\textcolor{red}{71.91}\tabularnewline

GFPGAN &	49.96 	&\textcolor{blue}{3.882}	&68.95	&111.78 	&\textcolor{blue}{4.349}	&70.58	&87.35 	&\textcolor{blue}{4.144}	&68.04	&77.52	&\textcolor{blue}{4.470}	&70.75\tabularnewline

DFDNet &	62.57     	&4.026	& 67.95	&111.55	&4.414	&69.44	&100.68	&5.293	&63.81 	&86.21	&4.679	&69.19\tabularnewline
\hline
DiffBIR &	\textcolor{blue}{38.97}	&5.351	&70.06	&116.56	&5.180	&68.26	&91.53	&5.744	&67.23	&91.90	&5.67	&68.75\tabularnewline

DiffMAC &	\textcolor{red}{37.75}	&5.777	&69.73	&120.17	&5.785	&68.09	&90.15	&5.741	&69.60	&96.82	&6.29	&69.22\tabularnewline
\hline
\end{tabular}
}
\label{tab:real}
\end{table*}

\begin{figure*}[htbp]
\begin{center}
   \includegraphics[width=0.78\linewidth]{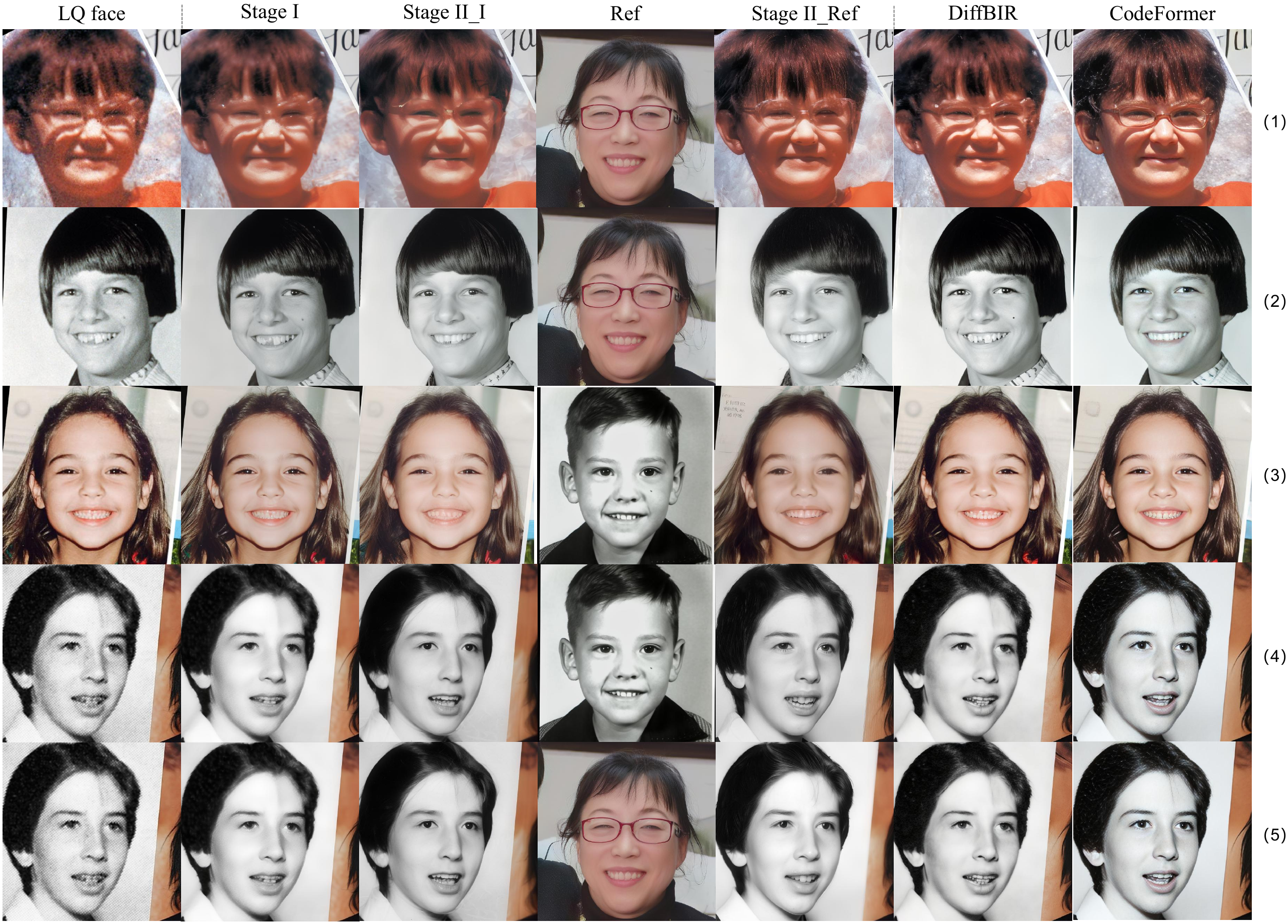}
\end{center}
   \caption{Original DiffMAC implements information compensation based on Stage I. Stage II equipped with id info of other reference faces (col 5) presents BFR with both high quality and fidelity as well, e.g., eyeglass (row 1), teeth (rows 2-5). Restoration diversity demonstrates the effectiveness of hallucination correction based on our proposed MIB. DiffBIR \cite{diffbir} and CodeFormer \cite{codeformer} generate unique BFR with topology or color hallucinations.}
\label{fig:diverse}
\end{figure*}

\textbf{CodeFormer} models the global face composition with codebook-level contextual attention, which implements stable BFR within the representation scope of the learned codebook. It is liable to fall into facial prior overfitting (col 2 in Fig \ref{fig:heter}), as well as the mismatch between predicted code and LQ input (row 4 in Fig \ref{fig:stage2}).

\textbf{VQFR}  is equipped with a VQ codebook dictionary and a parallel decoder for high-fidelity BFR with HQ textural details. It seems to synthesize sharper images with abundant textural details usually demonstrated by better NIQE \cite{NIQE} score (Tab \ref{tab:celeb}, Tab \ref{tab:real}). However, these detail enhancements are not conducted based on useful-information screening, which may have a serious impact on facial topology and global fidelity (col 7 in Fig \ref{fig:photo}).

\textbf{DifFace}  builds a Markov chain partially on the pre-trained diffusion reversion for BFR. Nevertheless, the intrinsic formulation is to approximate $q(x_{N}|x_{0})$ via $p(x_{N}|y_{0})$ where its diffused estimator and the fake $x_{N}$ may result in important information missing of original LQ face and uncontrollable diffusion reversion, illustrated by the distortions of glass, background, microphone, hand (col $3\&4\&5\&8$ in Fig \ref{fig:photo}). Furthermore,  the leveraged diffusion model was so overfitted on the FFHQ dataset that it synthesizes human-like faces in the heterogeneous domains (row 4 in Fig \ref{fig:heter}), although having better FID for HFW dataset (Tab \ref{tab:real}). Note that the measure anchor of FID is based on FFHQ.

\begin{figure*}[htbp]
\begin{center}
   \includegraphics[width=0.93\linewidth]{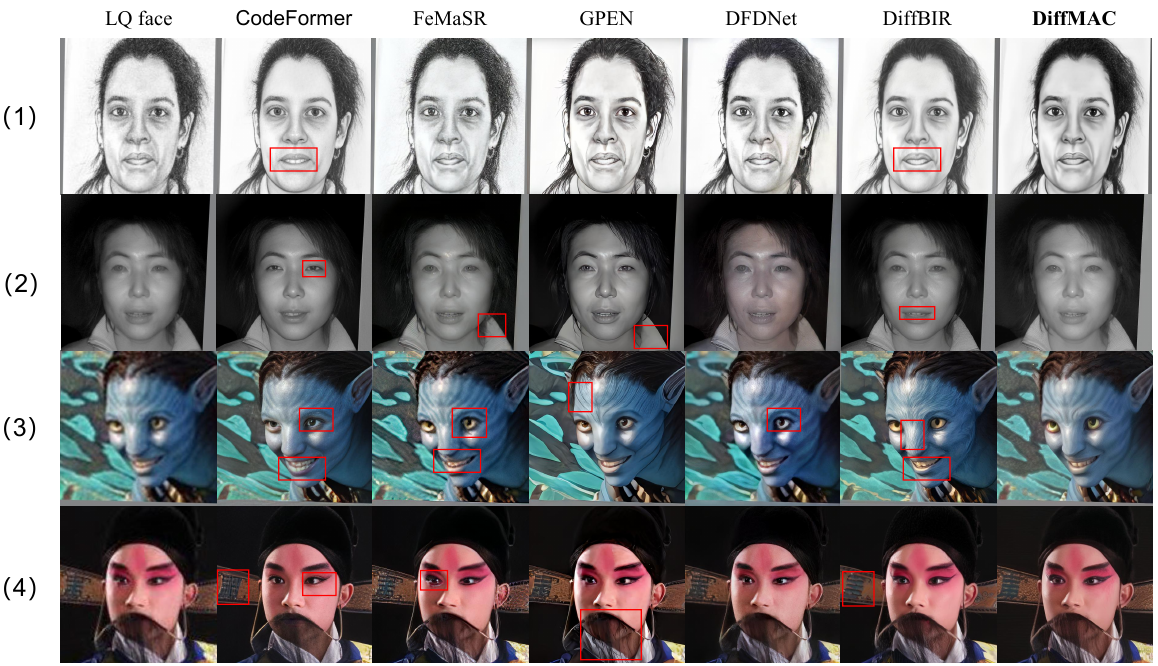}
\end{center}
   \caption{More qualitative results in heterogeneous scenes. CodeFormer \cite{codeformer} easily causes component mismatch, FeMaSR \cite{femasr} has messy and low-fidelity areas, GPEN \cite{GPEN} tends to synthesize sketchy contents, DFDNet \cite{dfdnet2020} also causes incorrect pattern matches, and DiffBIR \cite{diffbir} results in crucial detail missing and unfaithful texture diffusion. Nevertheless, DiffMAC respects heterogeneous styles and makes providential face restorations.}
\label{fig:heterm}
\end{figure*}

\textbf{FeMaSR} leverages the same learning paradigm as CodeFormer and VQFR, that is high-resolution prior storing along with feature matching. The training is conducted on patches to avoid content bias, which is beneficial to natural image restoration but not BFR with high facial composition prior in global view (Fig \ref{fig:photo}, Fig \ref{fig:heterm}).

\textbf{GPEN} first train StyleGAN from scratch with a modified GAN block, then embed multi-level features to the pre-trained prior model with concat operation. Similar to DifFace, GPEN employs latent codes from LQ face as the approximate Gaussian noises to modulate the pre-trained prior model, which is susceptible to intrinsic insufficient information of LQ face. Therefore, details are lacking on the restored face (col 3 in Fig \ref{fig:photo}, col 4 in Fig \ref{fig:heterm}), and even original hallucinations make face reconstruction deviate from the reasonable distribution (col 4 in Fig \ref{fig:photo}). 

\begin{figure}[htbp]
\begin{center}
   \includegraphics[width=1\linewidth]{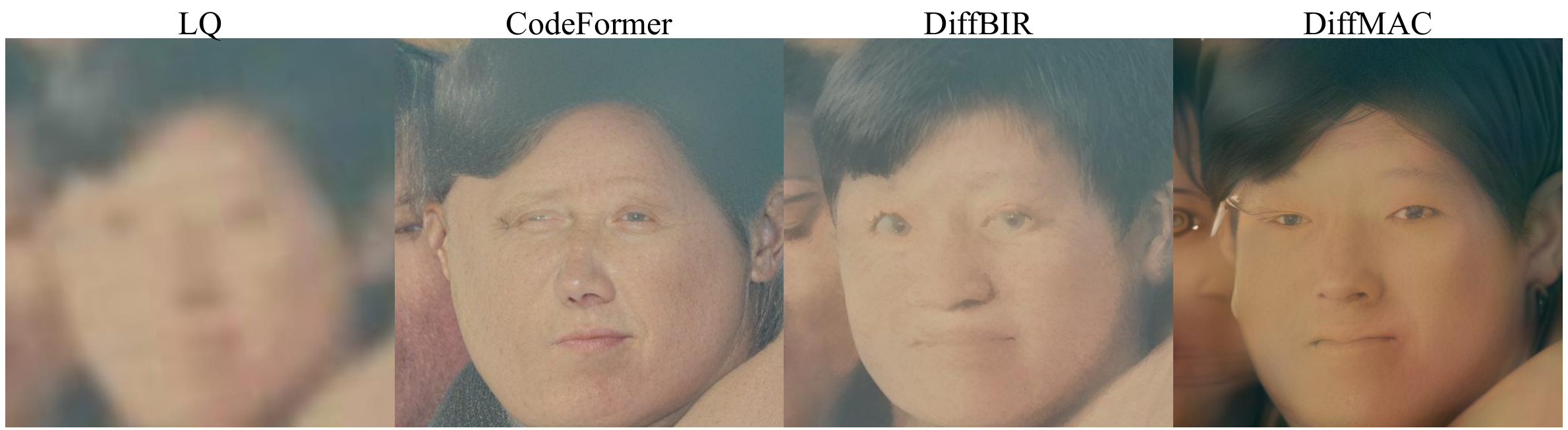}
\end{center}
   \caption{There are some failure generalization cases of DiffMAC in some severely degraded scenarios, which could be caused by the ambiguous facial contour.}
\label{fig:hard} 
\end{figure}

\textbf{GFPGAN} is another StyleGAN-prior-based method. Similar to GPEN, there is an uncontrollable risk while using the severely abstract latent from LQ face to drive the decoding of pre-trained StyleGAN. Moreover, the naive two-part channel splitting of the CS-SFT layer makes it hard to accurately incorporate realness-aware and fidelity-aware features, which usually results in attribute distortions, particularly in color space (Fig \ref{fig:photo}, Fig \ref{fig:heter}).

\textbf{DFDNet} conducts multi-level dictionary feature transfer based on HQ dictionaries of facial components. Nonetheless, it is challenging for out-of-component restoration (row 2 in Fig \ref{fig:heterm}) and precise HQ feature retrieval (row 3 in Fig \ref{fig:heterm}), especially in unknown and complicated degraded scenarios (Fig \ref{fig:photo}).

\textbf{DiffBIR} designs a two-stage framework to realize BFR including restoration module (RM) and generation module (GM) based on SwinIR and stable diffusion respectively. The results demonstrate RM produces an over-smoothed condition of GM (col 2 in Fig \ref{fig:photo}) and preserves hostile degradation deserving to be removed (col 7 in Fig \ref{fig:photo}). More examples are shown in Fig \ref{fig:stage2}.

Our DiffMAC not only leverages the strong generative ability of a stable diffusion model but also adopts information bottleneck to polish the hallucinated diffusion manifold, which is more adept at accommodating a diverse range of BFR scenarios (Fig \ref{fig:photo}, Fig \ref{fig:heter}, Fig \ref{fig:heterm}). 

\begin{table}[tbp]
\centering
\huge
\caption{Ablation study of DiffMAC variants on CelebChild. $no1$ means w/o Stgae I of DiffMAC, 2AdaIN indicates twice Stage I without MIB.}
\scalebox{0.38}{
\begin{tabular}{c|c|c|c}
\hline 
 Methods & FID$\downarrow$ & NIQE $\downarrow$ & MUSIQ$\uparrow$ \tabularnewline
\hline 
\hline
 
no1\_0.1\_n &122.53	&5.987	&66.74		\tabularnewline
no1\_0.01\_n&\textcolor{red}{115.17}	&5.481	&64.54	\tabularnewline
no1\_0.001\_n&116.83	&5.827	&65.93	\tabularnewline
0.1\_n&123.48	&5.649	&67.97	\tabularnewline
0.01\_n&120.65	&\textcolor{red}{5.332}	&68.10	\tabularnewline
0.001\_n&119.40	&5.696	&68.67	\tabularnewline
0.1\_id&125.91	&5.718	&\textcolor{red}{68.69}	\tabularnewline
0.01\_id&119.40	&5.477	&68.13	\tabularnewline
0.001\_id&120.17	&5.785	&68.09	\tabularnewline
stage I &117.27	&5.815	&65.79	\tabularnewline
MIB w/o n $\vert$ id &123.53	&5.628	&67.57	\tabularnewline
2AdaIN &117.94	&5.785	&67.63	\tabularnewline
\hline 
\end{tabular}}

\label{tab:ablation}
\end{table}
\begin{table}[htbp]
\centering
\huge
\caption{User study of DiffMAC and other state-of-the-art methods considering photorealistic and heterogeneous scenarios. Subjective metrics contain edge enhancement, color preservation of the LQ face, notable artifacts, heterogeneous style preservation, and overall subjective preference.}
\scalebox{0.38}{
\begin{tabular}{c|c|c|c|c|c}
\hline 
 Methods & Edge$\uparrow$ & Color$\downarrow$ & Artifacts $\downarrow$ &\makecell[c]{Hetero. \\style} $\uparrow$ & Preference$\uparrow$ \tabularnewline
\hline 
\hline
 
CodeFormer\cite{codeformer} &4.0	&1.8	&2.5	&2.5	&\textcolor{blue}{3.8}		\tabularnewline

VQFR\cite{vqfr} &3.7	&2.9	&3.6	&1.4	&3.0  	 \tabularnewline

DifFace\cite{difface} &2.9	&2.8	&4.3	&1.2	&2.3	\tabularnewline

FeMaSR\cite{femasr}&3.8	&\textcolor{blue}{1.4}	&4.2	&3.5	&2.2	\tabularnewline

GPEN\cite{GPEN} &	\textcolor{red}{4.3}	&\textcolor{red}{1.1}	&2.8	&\textcolor{blue}{4.0}	&3.1	\tabularnewline

GFPGAN\cite{gfp2021} &3.8	&4.6	&2.7	&1.1	&2.6	\tabularnewline

DFDNet\cite{dfdnet2020} &	3.1	&2.9	&\textcolor{blue}{2.0}	&2.9	&3.0	\tabularnewline

DiffBIR\cite{diffbir} &	2.2	&3.3	&4.8	&3.7	&2.9	\tabularnewline
\hline
DiffMAC &	\textcolor{blue}{4.1}	&1.6	&\textcolor{red}{1.2}	&\textcolor{red}{4.4}	&\textcolor{red}{4.3}\tabularnewline
\hline 
\end{tabular}}

\label{tab:human}
\end{table}

\subsection{Ablation Study}

We conduct qualitative and quantitative evaluation among different DiffMAC variants, as shown in Tab \ref{tab:ablation}, Fig \ref{fig:ablation}, and Fig \ref{fig:12}. 

$w/o\ Stage\ I$ is not friendly to conducting information compression-restoration trade-off, because a solid compression weight $\beta$ (e.g., 0.1, 0.01, 0.001) in Equ. \ref{equ:mib} needs to be explored for stable MIB during DiffMAC training. In general, lower-quality input is supposed to be imposed more information compression. Moreover, incautious MIB against unknown and complicated degradations on LQ faces achieves blurred faces. Note that only noise $\epsilon$ from $\mathcal{N}(\mu_{\mathcal{M}},\sigma_{\mathcal{M}}^{2})$ as information compensation to replace $\epsilon_{id}$ in Equ. \ref{equ:fuse}, because the destroyed identity information is ambiguous for LQ faces. Still and all, $no1\_0.01\_n$ gets good FID that demonstrates the effectiveness of MIB and AdaIN modulation.

$w/ noise\ injection, w/ Stage\ I$ is an alternative information compensation approach, but there may be uncontrollable artifacts (Fig \ref{fig:ablation}) introduced by random noise distribution, especially in the seriously distorted cases. Furthermore, while employing strong compression ($\beta=0.1$), it's hard for $w/noise\ injection$ to hold on original identity or expression compared with $w/id\ injection$ (Fig \ref{fig:12}).

$w/ id\ injection, w/ Stage\ I$ is capable of filtering stubborn restoration-irrelevant information (col 7 in Fig \ref{fig:photo}), meanwhile preserving the original identity, which demonstrates the essential role of MIB module with id info compensation. Moreover, different referenced identities produce diverse BFR results (Fig \ref{fig:diverse}).

Only $w/ Stage\ I$ more strictly respects the original content of the LQ face. However, It is still challenging for the model to defend severe degradations, and realize subjectively good BFR results (Fig \ref{fig:stage2}). This highlights the significance of information refinement based on MIB learning.

$MIB\ w/o\ n\vert id$ indicates only w/ information compressing, which will result in excessive information discarding demonstrated by distorted identity, expression, or textures (Fig \ref{fig:ablation}, Fig \ref{fig:12}). This highlights the important role of information compensation in MIB on precisely controllable BFR.

$2AdaIN$ is a naive two-stage variant without MIB. We find it's labored for this variant to remove restoration-irrelevant information while implementing uncontrollable texture syntheses and background restoration (Fig \ref{fig:motivation}). This provides evidence that MIB can achieve better learning outcomes.
\begin{table}[tbp]
\centering
\huge
\caption{Subjective study of Fig \ref{fig:motivation} about method preference.}
\scalebox{0.28}{
\begin{tabular}{c|c|c|c|c|c|c|c|c|c}
\hline 
 Methods & \makecell[c]{Code \\Former} & VQFR  & DifFace&FeMaSR& GPEN& GFPGAN& DFDNet & DiffBIR& DiffMAC \tabularnewline
\hline 
\hline
 
ph. : he.&\textcolor{blue}{9.0} : 1.9	&7.9 : 2.3	&5.5 : 0.8	&1.5 : \textcolor{red}{8.2}	&6.8 : 6.7	&6.3 : 1.5	&4.5 : 5.4	&7.9 : 3.8	&\textcolor{red}{9.2} : \textcolor{blue}{7.4}\tabularnewline
\hline 
\end{tabular}}

\label{tab:p2h}
\end{table}

\begin{table}[tbp]
\centering
\huge
\caption{Time study of DiffMAC and other state-of-the-art methods on NVIDIA RTX 3090. DifFace, DiffBIR, and DiffMAC take more time based on the diffusion model.}
\scalebox{0.3}{
\begin{tabular}{c|c|c|c|c|c|c|c|c|c}
\hline 
 Methods & \makecell[c]{Code \\Former} & VQFR  & DifFace&FeMaSR& GPEN& GFPGAN& DFDNet & DiffBIR& DiffMAC \tabularnewline
\hline 
\hline
 
Sec&0.028	&0.179	&4.146	&0.059	&0.052	&\textcolor{red}{0.028}	&0.682	&4.621	&10.780\tabularnewline
\hline 
\end{tabular}}

\label{tab:time}
\end{table}
\subsection{User Study}

Quantitative results (Tab \ref{tab:celeb}, Tab \ref{tab:real}, Tab \ref{tab:ablation}) are only the evaluation reference for performance, e.g., VQFR and DifFace get better scores in Tab \ref{tab:real}, but their visual BFR results seem messy in Fig \ref{fig:photo} and \ref{fig:heter}. 

User study plays an equally important role in the quality evaluation of HG-BFR tasks. We invite 10 users to conduct the subjective study. First, we briefly explain the BFR task and suggest users carefully observe the LQ and restored faces obtained by 8 state-of-the-art methods and our proposed algorithm. Each observed algorithm has 20 samples considering both photorealistic and heterogeneous domains. These observers need to give scores recorded as 1-5 from 5 aspects: (a) edge enhancement, (b) color distortion of the LQ face, (c) notable artifacts (dot, block, line, or uncomfortable artifacts perceived by human eyes, e.g., col $5\&2\&3$ in Fig \ref{fig:heter}, col 7 in Fig \ref{fig:photo}), (d) heterogeneous style preservation and (e) overall restored imaging preference, where the higher BFR quality is reflected by higher scores of (a), (d), (e) as well as lower scores of (b), (c). 

Finally, we collect 200 score tables, where each table contains 45 human decisions corresponding to different methods and subjective metrics. The statistic average scores are recorded in Tab \ref{tab:human}. DiffMAC presents a more competitive human-perceptual performance. Furthermore, we show the subjective study of Fig \ref{fig:motivation} in 
Tab \ref{tab:p2h} where we collect 10 score tables for the photo. : hetero., which demonstrates the high-quality BFR capability of DiffMAC concerning both photorealistic and heterogeneous domains. 

\subsection{Limitations}

It is still challenging for DiffMAC to handle BFR in some unseen scenarios (Fig \ref{fig:hard}). As for inference time, It takes more time for DiffMAC compared to DiffBIR due to the conduction of Stage I and Stage II with MIB (Tab \ref{tab:time}). Future work will explore the efficient distillation of the DDIM sampling. 

%


\section{Conclusion}
We first study diffusion manifold hallucination correction for high-generalization BFR. It is meaningful to explore information restoration based on information compressing and information compensation on manifold space. We propose a diffusion-information-diffusion framework for high-fidelity as well as high-quality blind face restoration. Extensive experimental analyses have shown the superiority of DiffMAC in complicated wild and out-of-distribution heterogeneous scenarios.





%

\end{document}